\documentclass[lettersize,journal]{IEEEtran}
\usepackage{amsmath,amsfonts}
\usepackage{algorithmic}
\usepackage{algorithm}
\usepackage{array}
\usepackage[caption=false,font=normalsize,labelfont=sf,textfont=sf]{subfig}
\usepackage{textcomp}
\usepackage{booktabs}
\usepackage{array,multirow,tabularx}
\usepackage{stfloats}
\usepackage{url}
\usepackage{verbatim}
\usepackage{graphicx}
\usepackage{cite}
\usepackage[colorlinks,citecolor=blue]{hyperref}
\begin{document}

\title{Combining Self-attention and Dilation Convolutional for Semantic Segmentation of Coal Maceral Groups}

\author{{Zhenghao Xi,~\IEEEmembership{Member,~IEEE}, Zhengnan Lv,Yang Zheng, ~\IEEEmembership{Member,~IEEE}, Xiang Liu,~\IEEEmembership{Member,~IEEE}, Zhuang Yu ,Junran Chen, Jing Hu and Yaqi Liu }
	\thanks{Manuscript received **, **; revised ** **, **.(\textsl{Corresponding author:Zhengnan Lv)}}
	\thanks{Zhenghao Xi, Xiang Liu Junran Chen and Yaqi Liu are with the School of Electronic and Electrical Engineering, Shanghai University of Engineering Science, Shanghai 201620, China. (e-mail: zhenghaoxi@hotmail.com;  xliu@sues.edu.cn; Junran0402@163.com; ulrich@163.com)}
		\thanks{Zhengnan Lv and Jing Hu are with China Electronics Technology Group Corporation No.15 Research Institute,Beijing 100083,China (e-mail: lvzhn0@163.com; hjing1126@163.com).}
		\thanks{Yang Zheng  is with the Institute of Automation, Chinese Academy of	Sciences, Beijing 100101, China (e-mail: yang.zheng@ia.ac.cn)}
    	\thanks{Zhuang Yu is with the Department of Manufacture Management of Angang Steel Co.,Ltd., Liaoning 114031, China (e-mail: 619330252@qq.com)}}

\maketitle

\begin{abstract}
The segmentation of coal maceral groups can be described as a semantic segmentation process of coal maceral group images, which is of great significance for studying the chemical properties of coal. Generally, existing semantic segmentation models of coal maceral groups use the method of stacking parameters to achieve higher accuracy. It leads to increased computational requirements and impacts model training efficiency. At the same time, due to the professionalism and diversity of coal maceral group images sampling, obtaining the number of samples for model training requires a long time and professional personnel operation. To address these issues, We have innovatively developed an IoT-based DA-VIT parallel network model. By utilizing this model, we can continuously broaden  the dataset through IoT and achieving sustained improvement in the accuracy of coal maceral groups segmentation. Besides, we decouple the parallel network from the backbone network to ensure the normal using of the backbone network during model data updates. Secondly, DCSA mechanism of DA-VIT is introduced to enhance the local feature information of coal microscopic images. This DCSA can decompose the large kernels of convolutional attention into multiple scales and reduce 81.18\% of parameters.Finally, we performed the contrast experiment and ablation experiment between DA-VIT and state-of-the-art methods at lots of evaluation metrics. Experimental results show that DA-VIT-Base achieves 92.14\% pixel accuracy and 63.18\% mIoU. Params and FLOPs of DA-VIT-Tiny are 4.95M and 8.99G, respectively. All of the evaluation metrics of the proposed DA-VIT are better than other state-of-the-art methods.
\end{abstract}


\section{Introduction}
\IEEEPARstart{C}{oal}, as an important fossil energy, plays a crucial role not only in addressing energy demands but also in various fields such as power generation, coking, and chemical industries \cite{tang2020research,li2022coal,liu2022role}. It provides essential raw materials for daily consumer goods and remains the core of the energy system in the present and future \cite{dai2018coal,tong2022coal}. In order to distinguish the quality of coal and improve its utilization efficiency, the analysis of vitrinite reflectance and coal maceral groups is performed \cite{hou2020differences}, which allows for the classification and evaluation of coal properties and applications. At the same time, the environ12 mental problems caused by the fossil energy are becoming even more severe. In order to achieve the protection of the ecological environment, as well as Chinese carbon peak and carbon neutral plan, it is particularly necessary to do a good job in the separation and precision utilization of macerals in coal \cite{mao2022insights,mollo2022simultaneous}, and the analysis of coal maceral groups plays an important role \cite{hou2020differences,burnham2019kinetic,hower2022understanding}.

The complexity and significant variations are the characteristics of the coal maceral groups \cite{wang2022exploring}, moreover, the semantic segmentation and analysis of maceral groups mainly rely on manual efforts, which is subject to the experience and subjectivity of the operators\cite{yuan2024secure}. Thus, the manual efforts consume a considerable amount of time. Meanwhile, there are differences between the test results of the operator, the uniformity cannot be guaranteed. In contrast, image analysis technology using computers greatly improves the accuracy of detection and the efficiency of analysis in actual situations \cite{cao2021toward},\cite{cao2022blockchain}.

Many methods based on machine learning and image processing have been applied to maceral group analysis yielding remarkable results \cite{wang2019intelligent}. Wang et al. constructed an initial feature set by extracting features such as the gray level cooccurrence matrix from coal images \cite{wang2017classification}. Then, they reduced the dimensionality using PCA, and classified the vitrinite using the SVM algorithm. Khandelwal et al. estimated the composition of Indian coal by designing an expert system, that combines the imperialist competitive algorithm and artificial neural networks \cite{khandelwal2017expert}. Song et al. proposed a method to solve the problem of false boundary image elimination using Prewitt operator for coal maceral groups segmentation, and they used the K-means clustering algorithm for coal maceral groups recognition \cite{song2019automatic,song2019effect}. Onifade et al. analyzed the coal maceral groups in samples from the Witbank coalfield using multiple-input single-output white box artificial neural networks, multivariate linear regression, and multivariate nonlinear regression methods, moreover, they explored the differences and efficiencies among these methods for coal maceral groups segmentation \cite{onifade2021development}. However,these traditional machine learning approaches require the substantial amount of domain knowledge and experience \cite{malik2020hierarchya}\cite{roscher2020explainable}, which not only leads to models often having high computational complexity, but also requires time-consuming and laborious professional annotation of a large number of samples.

The main contributions of this paper are as follows:

(1) We constructed a novel Dilation-based Convolutional Self-Attention (DCSA) mechanism constructed with a new coal maceral groups segmentation model based on the industrial internet of things and called Dilated Attention Vision Transformer (DA-VIT) using the DCSA is proposed in this paper for extracting features of coal maceral groups. Moreover, the proposed model preserves the adaptability of coal microscopic images in the channel dimension, avoiding the issue of parameter explosion caused by large convolutional kernels. Furthermore, the proposed DA-VIT improves accuracy and reduces floating-point operations per second (FLOPs) compared with other state-of-the-art methods.

(2) We have innovatively designed an IoT-based DA-VIT parallel network model. By utilizing this model, we can continuously broaden the dataset through IoT and achieving sustained improvement in the accuracy of coal maceral groups segmentation. We decouple the parallel network from the backbone network to ensure the normal using of the backbone network during model data updates.

\section{Related Work}
In recent years, the IoT has made important progress in the research and practice of smart medical and industrial IoT \cite{kopetz2022internet}, with the help of data collection and remote control of multi-terminal devices in the IoT, the efficiency of medical treatment and industrial production has been greatly improved, showing the great potential of IoT technology in practical applications. Research on new architectures such as edge computing \cite{cao2020overview} and fog computing \cite{yi2015survey} has gradually become an important approach to the data acquisition of IoT terminals, which not only reduces the dependence on the central server by the terminal equipment processes the data, but also reduces the delay of data transmission.

To overcome the aforementioned issue, various studies on the semantic segmentation of coal maceral groups have used deep learning methods. U-Net is one of them with the main focus that has a U-shaped architecture and skip connections, which ensures segmentation accuracy with less training data \cite{ronneberger2015uneta}. Lei et al. improved the U-Net by integrating Attention Gate \cite{vaswani2017attention,lei2021maceral}. This method suppressed irrelevant information and noise and further improved the segmentation of coal maceral groups. Fan et al. studied the various U-Net variants, and investigated the particle morphology, grain size, dissociation characteristics, and density separation processes of coal maceral groups using each of them \cite{fan2023macerals}. However, the optimization models based on U-Net have limitations with fixed receptive fields, which leads to unsatisfactory effects in handling high-resolution coal microscopic images. To address this issue, the Deeplabv3+ \cite{chen2018encoderdecoder} model was used to solve the coal maceral groups segmentation tasks \cite{wang2019intelligent}. This model expanded its receptive field on coal microscopic images by the dilated convolutions and improved the segmentation accuracy. On the other hand, Deeplabv3+ generates high computational complexity for capturing more contextual information, this deficiency restricts the application of the Deeplabv3+ to high-resolution coal microscopic image processing. 

In recent years, Liu et al. designed a Swin-Transformer deep learning model based on VIT model \cite{dosovitskiy2021image}, which can handle high-resolution images by utilizing the shifted windows. Luan et al. solved the image classification problem of coal and coal gangue using the Swin-Transformer \cite{luan2023coal}, of which experimental results showed that the Swin-Transformer enhanced the learning and processing capabilities of high-resolution coal microscopic images. However, the self-attention mechanism of the Swin-Transformer model will convert the coal maceral groups image into a one-dimensional sequence when the image preprocessing, this results in the loss of the two-dimensional spatial features of coal maceral groups. To address the problem, researchers proposed using convolutional self-attention with a large kernel, and used it to replace the traditional one \cite{woo2018cbam,hu2018gatherexcite}. This approach can preserve the long-range dependencies between pixels in coal maceral groups while retaining the local feature information extracted by convolution.

\section{Our Approachs}
In this section, we first research the feature extraction of coal microscopic images using dilated convolution. Then, we introduce the attention mechanism into dilated convolution to address the issue of weak correlations between coal maceral groups, thereby developing the Dilation-based Convolutional Self-Attention (DCSA). Finally, we construct a new segmentation model for coal maceral groups based on the industrial IoT, which we call the Dilated Attention Vision Transformer (DA-VIT), utilizing the DCSA for feature extraction of coal maceral groups.
\subsection{Feature extraction of coal microscopic images based on dilated convolution}
The deep learning model has become the mainstream approach for solving the semantic segmentation task of coal microscopic images \cite{lei2021maceral,fan2023macerals,wang2022identification}. The model uses traditional convolutional operations to compress the feature information of coal microscopic images. Through continuous down-sampling, the convolutional operations reduce the dependency of the model on detailed information and achieve stronger receptive field \cite{dong2022remote}. Thus, the deep learning model can understand the coal microscopic images deeper and improve the accuracy of image segmentation.

However, when we extract the feature of neighboring pixels, traditional convolutional operations often result in overlap calculations. It leads to a reduction in the computational efficiency of the deep learning model.

\begin{figure}[!ht]
	\centering
	\includegraphics[width=2.5in]{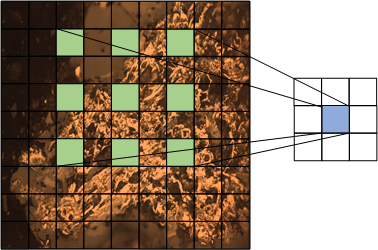}
	\caption{Diagram of dilated convolution.}
	\label{fig_1}
\end{figure}

The dilated convolution \cite{yu2017dilated} is proposed to solve this problem. Fig. \ref{fig_1} illustrates the diagram of dilated convolution with $r$=2, where $r$ is a dilation rate within the convolution kernel. As shown in the Fig. \ref{fig_1}, the dilated convolution can extract features from a larger range. It mitigates the impact of complexity and significant variations \cite{wang2022exploring} between coal maceral groups. To capture features from different ranges, we mix the dilated convolutions with different $r$. It is especially suitable for coal microscopic images where pixels have great influence on the surrounding area.

\subsection{Dilation-based convolutional self-attention (DCSA)}
In order to solve the problem of the correlations between coal maceral groups are weak when we use dilated convolution, we introduce the attention mechanism and propose the Dilation-based Convolutional Self-Attention (DCSA) in this section.

There are various factors can influence on the structures of different maceral groups in coal microscopic images, such as position, temperature, mining, and processing. The dilated convolution operations focus on local information, lacking unified features, and cannot get the correlations between coal maceral groups. In order to improve the segmentation effect of the model, we need to extract the feature information on a wider range and collect long-range dependencies to establish the relationship between different pixels of coal microscopic images. Therefore, we introduce the attention mechanism in this paper as the strategy.

\begin{figure}[!ht]
	\centering
	\includegraphics[width=2.5in]{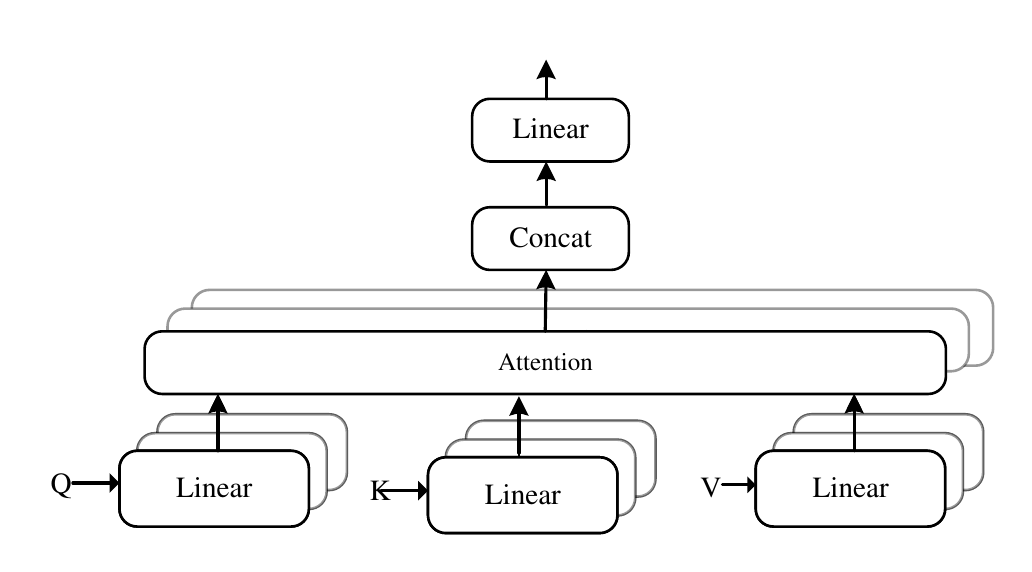}
	\caption{Attention mechanism.}
	\label{fig_2}
\end{figure}

The attention mechanism maps image features to multiple sets of $Q$, $K$, $V$ (Query, Key, and Value). By computing multiple attention weights in parallel and then concatenating them, the attention mechanism captures information from different aspects of the input. It strengthens the connectivity between pixels in different regions of coal microscopic images. The attention mechanism is illustrated in Fig. \ref{fig_2}. The formula of the attention mechanism is shown in Eq. 1.

\begin{equation}
\label{deqn_ex1a}
{\mathop{\rm Attention}\nolimits} (Q,K,V) = {\mathop{\rm softmax}\nolimits} (\frac{{Q{K^T}}}{{\sqrt {{d_k}} }})V
\end{equation}
where $d_k$ is the dimension of the $Q$.

However, the traditional image attention mechanism neglects the two-dimensional information of the image and fails to adapt to the channel dimension.

Another approach is convolutional self-attention with large kernels \cite{woo2018cbam,hu2018gatherexcite}. However, the size of the convolutional kernel is related to the segmentation accuracy, which requires a large number of parameters.

\begin{figure}[!ht]
	\centering
	\includegraphics[width=2.5in]{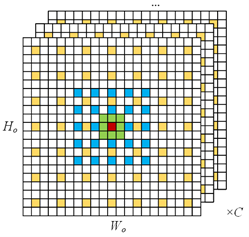}
	\caption{Illustration of DCSA.}
	\label{fig_3}
\end{figure}

To reduce parameters, computation, and hardware requirements while ensuring segmentation accuracy, we propose the dilation-based convolutional self-attention (DCSA) mechanism by integrating convolutional image attention with dilated convolutions at different $r$, as shown in Fig. \ref{fig_3}. This approach preserves the long-range dependencies captured by self-attention while enhancing the acquisition of two-dimensional information from the image. Moreover, the introduced method enables the retrieval of long-range information from different ranges and combines the information to compute the attention map.

\begin{figure}[!ht]
	\centering
	\includegraphics[width=\linewidth]{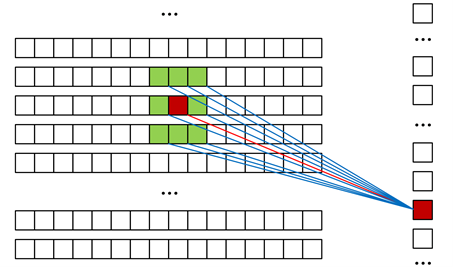}
	\caption{SI-Conv diagram.}
	\label{fig_4}
\end{figure}

The large convolutional kernel is decomposed into three parts as indicated in Fig. \ref{fig_3}. The short-range information convolution (SI-Conv), medium-range dilation convolution (MD-Conv), and long-range dilation convolution (LD-Conv) are described in green, blue, and yellow, respectively. Through DCSA, we can preserve the multiple directions of continuous spatial information. Moreover, the SI-Conv is illustrated in Fig. \ref{fig_4}. MD-Conv and LD-Conv can be obtained by changing the kernel size and dilation rate $r$. Their structures are similar to SI-Conv.

For a $K \times K$ dilated convolution, the equivalent kernel size of its receptive field is $K' = K + (K - 1) \times (r - 1)$. The reduction rate of parameters can be calculated using Eq. 1.

\begin{equation}
	\label{deqn_ex1a}
\rho  = 1 - \frac{{\sum\limits_{i = 3}^3 {{K_m} \times {K_m} \times C} }}{{{H_0} \times {W_0} \times C}}
\end{equation}
where $K_m$ is the different scales of decomposition. $C$ is the number of channels. And $H_o$ and $W_o$ represent the size of the large kernel. The size of kernel is ${\rm{19}} \times {\rm{19}} \times {\rm{C}}$. we decomposed it into a ${\rm{3}} \times {\rm{3}} \times {\rm{C}}$ SI-Conv, a ${\rm{5}} \times {\rm{5}} \times {\rm{C}}$ MD-Conv, and a ${\rm{7}} \times {\rm{7}} \times {\rm{C}}$ LD-Conv. The reason is the contribution of long-range to accuracy decreases when the distance between pixels increases \cite{li2023uniformer}. The $\rho$  of DCSA is 81.18\%.

\begin{figure}[!ht]
	\centering
	\includegraphics[width=\linewidth]{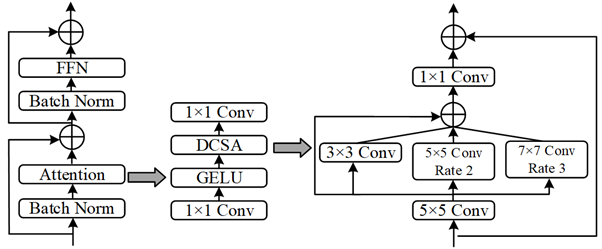}
	\caption{Schematic diagram of the attention module.}
	\label{fig_5}
\end{figure}

The module to extract feature information of coal maceral groups is depicted in Fig. \ref{fig_5}. It first extracts local information and then performs multi-scale feature extraction through multi-scale dilated convolutions in Fig. \ref{fig_3}. Finally, it combines the information from different channels through a ${\rm{1}} \times {\rm{1}}$ convolution to enhance the adaptability of the model in the channel dimension. The resulting attention map is obtained as the output. The expression for the attention map Att computed by DCSA is given by Eq. 3. The result obtained by weighting the input according to Att is expressed by Eq. 4.
\begin{equation}
	\label{deqn_ex1a}
	\begin{aligned}
		Att = &{{\mathop{\rm Conv}\nolimits} _{1 \times 1}}(\sum\limits_{s = 3,5,7}^{r = 1,2,3} {{{{\mathop{\rm Conv}\nolimits} }_{s \times s}}({{{\mathop{\rm Conv}\nolimits} }_{5 \times 5}}(Input),r)}\\
		&  + {{\mathop{\rm Conv}\nolimits} _{5 \times 5}}(Input))
	\end{aligned}
\end{equation}
\begin{equation}
	Output = Att \otimes Input\label{deqn_ex1b}
\end{equation}
where ${{{{\mathop{\rm Conv}\nolimits} }_{s \times s}}}$ is a convolution with a kernel size of $s \times s$.$r$ s the dilation rate. And $ \otimes $ denotes element-wise multiplication of matrices.

\subsection{The DA-VIT parallel network model based on IoT}
To avoid the problem of network training models being unusable due to dataset updates, we have established the DA-VIT parallel network model based on IoT in this section. Our coal microscopic images have a large resolution which is 2048×1536, which makes it difficult to capture the information of the overall image when we segment the coal maceral groups at a single scale, and the computational requirements have also been increased. Moreover, the correlation among the morphological features of different coal microscopic images is weak low. A single-scale model structure makes it very hard to focus on detailed shapes and the overall construction of coal maceral groups simultaneously. All of these can affect the segmentation accuracy.

Our coal microscopic images have a large resolution which is ${\rm{2048}} \times 1536$, which makes it difficult to capture the information of the overall image when we segment the coal maceral groups at a single scale, and the computational requirements have also been increased. Moreover, the correlation among the morphological features of different coal microscopic images is weak low. A single-scale model structure makes it very hard to focus on detailed shapes and the overall construction of coal maceral groups simultaneously. All of these can affect the segmentation accuracy.

Obtaining coal maceral groups images is time-consuming and labor-intensive, and we require the model to fully learn the features of each component in each image.Therefore, we develop a DA-VIT parallel network model based on IoT, as shown in Fig. \ref{fig_6}.

\subsubsection{The construction of DA-VIT}
\begin{figure}[!t]
	\centering
	\includegraphics[width=\linewidth]{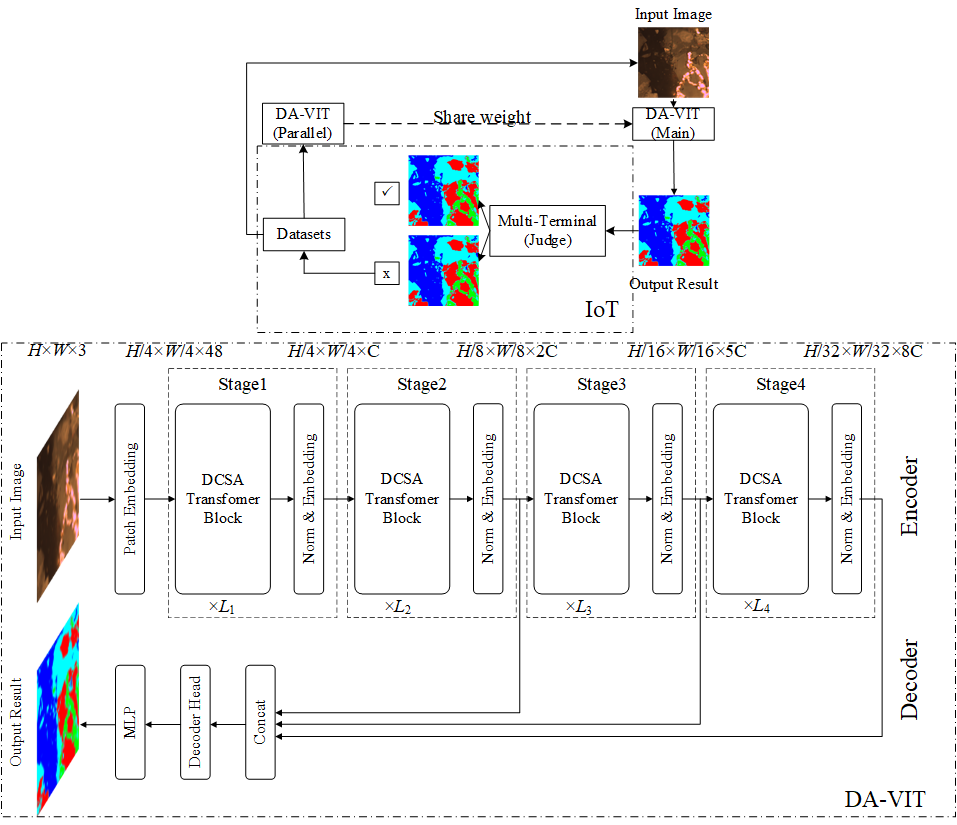}
	\caption{Illustration of DA-VIT Structure}
	\label{fig_6}
\end{figure}

We propose the multi-scale DA-VIT model by integrating DCSA into the framework structure of the Swin-Transformer \cite{liu2021swin}. The DA-VIT Structure is shown in Fig. \ref{fig_6}, where $C$ is the channel dimension, and $L_n$($n$=1,2,3,4) represents the number of DCSA blocks in each stage. Moreover, the DA-VIT model focuses on features at different levels of coal microscopic images by the multi-scale model structure of Swin-Transformer, and it can capture the dependency information of these features at different scales using leverages DCSA. 

In the encoder of DA-VIT, we adopt the four-layer pyramid structure that is similar to the Swin-Transformer. Stages 1 and 2 extract the features such as boundaries, textures, and shapes of coal maceral groups. Stages 3 and 4 extract the higher-level features such as overall structures and semantic information of coal microscopic images. In every stage, DCSA captures the dependency relationships between pixels at different ranges in the coal microscopic images. In our pyramid structure, the resolution decreases layer by layer, which are $H/4 \times W/4$, $H/8 \times W/48$, $H/16 \times W/16$, and $H/32 \times W/32$, respectively. $H$ and $W$ are the height and width of the coal microscopic images. The $C$ is increased with the image resolution decreases.

In the decoder of DA-VIT, we adopt the lightweight decoding head \cite{geng2021attention}, which can effectively improve the performance and efficiency of the model \cite{chen2018encoderdecoder,li2023uniformer,xie2021segformer,guo2022segnext}. Additionally, skip connections are added in our decoder to enhance the decoding effect by integrating the feature information from Stage 2 to Stage 4.

n different Stage or model sizes, the $C$ and $L_n$ which is the number of DCSA in each Stage are different. The values of $C$ and $L_n$ ($n$=1, 2, 3, 4) in Table \ref{tab1}.

\begin{table}[!ht]
	\centering
	\caption{Setting parameters for different model scale}
	\label{tab1}
	\begin{tabular}{cccc}
		\toprule
		Stage & Tiny & Small & Base \\
		\midrule
		C & 32 & 64 & 64 \\
		L1 & 3 & 2 & 3 \\
		L2 & 3 & 2 & 3 \\
		L3 & 5 & 4 & 12 \\
		L4 & 2 & 2 & 3 \\
		Params(M) & 4.95 & 14.74 & 27.57 \\
		\bottomrule
	\end{tabular}
\end{table}

\subsubsection{The design of the DA-VIT structure in IoT}
DA-VIT uses effective data obtained from IoT to update the weights of the parallel network generally, but this process will disturb the use of it. We developed the DA-VIT parallel network model to address this problem, as shown in Fig. \ref{fig_6}, where both the backbone network and the parallel network are DA-VIT models as described in Section \uppercase\expandafter{\romannumeral3}. C . 1)

In the DA-VIT parallel network model based on IoT, we send coal maceral groups images obtained from IoT user terminal into the backbone network for prediction. Experts judge the predicted results and then evaluate it for quality. We add unqualified results to the dataset to train the parallel network, along with their corresponding original images and labels. This process improves the segmentation accuracy of DA-VIT for coal maceral groups under various conditions. We share parallel network weights with the backbone network for prediction  when the parallel network training is complete.

In the DA-VIT parallel network module, data updates in the parallel network won’t affect the usage of the backbone network. It means during the training of the parallel network; users can still use the backbone network to predict input images and evaluate segmentation results. This approach meets the industrialization requirements of the model and ensures the efficiency of dataset expansion.

Besides, the proposed DA-VIT parallel network model based on IoT is also a closed-loop model. During the evaluation of predicted image results using expert judgment, the model continuously expands the dataset with images of coal maceral groups that DA-VIT has not fully learned. It enhances the segmentation accuracy of DA-VIT.

\section{Experiment}
\subsection{Experiment dataset based on the industrial IoT}
The image samples of coal maceral groups used in the experiment is collected from 12 observation terminal in the industrial internet of things by the collaborative institutions, and the samples have different coal samples with varying degrees of metamorphism. So, the experimental dataset is more comprehensive, which also makes our method more universal.

In order to obtain a large number of multi-source training images of coal maceral groups in a short period of time, we innovatively use the IoT to solve this problem in this paper. Firstly, we create a dedicated IoT which consists of the 7 computer terminals used for coal maceral group segmentation at different subsidiaries of our cooperative units, and deploy simple gamma correction and histogram equalization image processing algorithms on each terminal. We obtain 700 coal microscopic group images in just 5 days, and these images come from different terminals and different types of coal, which also makes the data trained by the model more generalizable. Then, we process and select the 700 coal microscopic group images using the method of \cite{hu2023improved}, and finally obtain 70,000 coal microscopic group image samples with the spatial shape of 512*512. We construct the dataset used 70,000 coal microscopic group image samples.

Moreover, we will continuously obtain new segmentation result data by deploying the segmentation method obtained from training the dataset on each terminal of the IoT network. The new segmentation result data continue to supplement our dataset to optimize the model training results. We achieve a closed-loop from "data acquisition", "data processing", to "result sending" to solve the problems of high time cost, slow inference speed, and low segmentation accuracy in the current method using deep learning models to process coal maceral group segmentation.

\begin{figure}[!ht]
	\centering
	\includegraphics[width=\linewidth]{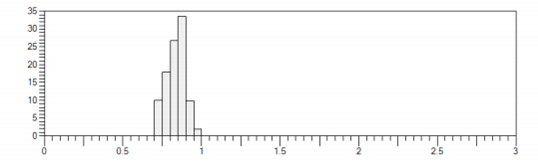}
	\caption{Vitrinite reflectance distributions of Coal A}
	\label{fig_7}
\end{figure}

\begin{figure}[!ht]
	\centering
	\includegraphics[width=\linewidth]{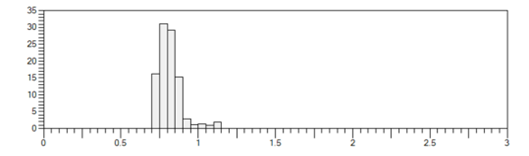}
	\caption{Vitrinite reflectance distributions of Coal B}
	\label{fig_8}
\end{figure}

\begin{figure}[!ht]
	\centering
	\includegraphics[width=\linewidth]{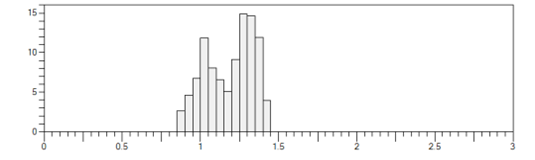}
	\caption{Vitrinite reflectance distributions of Coal C}
	\label{fig_9}
\end{figure}

\begin{figure}[!ht]
	\centering
	\includegraphics[width=\linewidth]{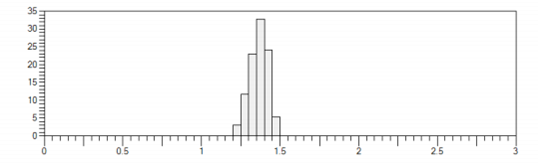}
	\caption{Vitrinite reflectance distributions of Coal D}
	\label{fig_10}
\end{figure}

\begin{figure}[!ht]
	\centering
	\includegraphics[width=\linewidth]{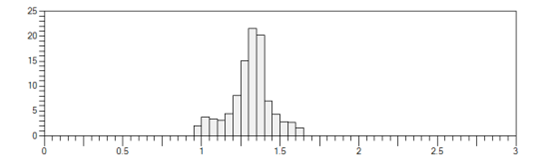}
	\caption{Vitrinite reflectance distributions of Coal E}
	\label{fig_11}
\end{figure}

\begin{figure}[!ht]
	\centering
	\includegraphics[width=\linewidth]{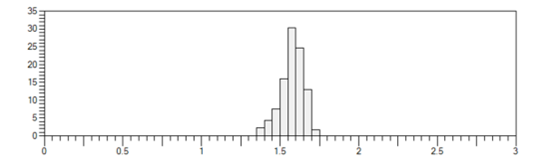}
	\caption{Vitrinite reflectance distributions of Coal F}
	\label{fig_12}
\end{figure}

\begin{figure}[!ht]
	\centering
	\includegraphics[width=\linewidth]{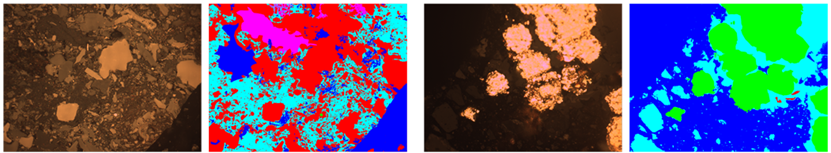}
	\caption{Microscopic images of coal and labeling example}
	\label{fig_13}
\end{figure}

There are 79 images of coal maceral groups in our experimental dataset. Coal routine test results are shown in Table \ref{tab2}. Three coal experts performed pixel-level annotations on all of the image samples. The image samples collection and preparation process followed the ISO18283:2006 standard. Vitrinite reflectance ranges from 0.61\% to 1.70\%. Vitrinite reflectance distribution is shown in Table \ref{tab3}. Different Coal Vitrinite reflectance distributions are shown in from Fig. \ref{fig_7} to Fig. \ref{fig_12}. The images were obtained using a Zeiss Axioskop40 optical microscope with a yellow filter magnified 500 times, and had a resolution of 2048×1536 and RGB format. According to the classification standard of coal maceral groups ISO7404-2, the maceral groups were categorized into vitrinite, inertinite, exinite, and mineral. The partial image samples of our dataset are shown in Fig. \ref{fig_13}, where light blue represents the vitrinite, red represents the inertinite, pink represents the exinite, green represents the mineral, and dark blue represents the adhesive.

\begin{table}[!ht]
	\centering
	\caption{Coal routine test results}
	\label{tab2}
	\begin{tabular}{cccccc}
		\toprule
		\multirow[m]{2}{*}{Coal code} & \multirow[m]{2}{*}{Coal grade} & \multicolumn{3}{c}{technic index}  & \multirow[m]{2}{*}{$R_{\max}^0$/\%}   \\
		\cmidrule{3-5}
		&  & $V_{daf}$ /\% & $Y/\mathrm{mm}$ & G & \\
		\midrule
		A & QM & 36.35 & 8.5 & 59 & 0.889   \\
		B & 1/3JM & 37.44 & 14 & 86 & 0.868   \\
		C & FM & 33.15 & 24 & 93 & 1.264   \\
		D & JM & 27.46 & 20.5 & 88 & 1.453   \\
		E & JM & 23.78 & 19 & 81 & 1.395   \\
		F & SM & 18.31 & 21 & 25 & 1.678   \\
		\bottomrule
	\end{tabular}
\end{table}

\begin{table}[!ht]
	\centering
	\caption{Composition analysis of coal maceral groups}
	\label{tab3}
	\tabcolsep5pt
	\extrarowheight2pt
	\aboverulesep0pt
	\belowrulesep0pt
	\begin{tabular}{ccccccc}
		\toprule
		\multirow[m]{2}{*}{Reflectance interval}
		& \multicolumn{6}{c}{Frequency distribution of vitrinite (\%)}  \\
		\cmidrule{2-7}
		& A & B & C & D & E & F \\
		\midrule
		$<$0.5 & 0 & 0 & 0 & 0 & 0 & 0 \\
		0.5-0.6 & 10.27 & 0 & 0 & 0 & 0 & 0 \\
		0.6-0.7 & 10.27 & 47.31 & 0 & 0 & 0 & 0 \\
		0.7-0.8 & 78.15 & 47.31 & 2.65 & 0 & 0 & 0 \\
		0.8-0.9 & 78.15 & 44.63 & 2.65 & 0 & 0 & 0 \\
		0.9-1.0 & 11.58 & 8.06 & 35.2 & 0 & 10.64 & 0 \\
		1.0-1.1 & 11.58 & 8.06 & 35.2 & 0 & 10.64 & 0 \\
		1.1-1.2 & 0 & 8.06 & 62.15 & 0 & 0 & 0 \\
		1.2-1.3 & 0 & 0 & 62.15 & 100 & 0 & 0 \\
		1.3-1.4 & 0 & 0 & 62.15 & 100 & 0 & 0 \\
		1.4-1.5 & 0 & 0 & 62.15 & 100 & 0 & 0 \\
		1.5-1.6 & 0 & 0 & 0 & 0 & 5.15 & 75.93 \\
		1.6-1.7 & 0 & 0 & 0 & 0 & 5.15 & 75.93 \\
		1.7-1.8 & 0 & 0 & 0 & 0 & 0 & 1.77 \\
		1.8-1.9 & 0 & 0 & 0 & 0 & 0 & 1.77 \\
		1.9-2.5 & 0 & 0 & 0 & 0 & 0 & 0 \\
		$\geq$2.5 & 0 & 0 & 0 & 0 & 0 & 0 \\	
		\bottomrule
	\end{tabular}
\end{table}

\subsection{Experimental data preprocessing}
Deep learning models suffer from overfitting or underfitting problems \cite{jabbar2014methods}. The main reason is insufficient training data and low model complexity. To avoid such problems, we need to preprocess the image samples of coal maceral groups from the experimental dataset of Section \uppercase\expandafter{\romannumeral3}. A. In every preprocessing step, there are multiple enhanced data samples have been generated. Finally, we expand the number of image samples up to 7900 from the experimental dataset of Section \uppercase\expandafter{\romannumeral4}. A. The specific preprocessing steps are as follows:

(1) We randomly cropped the original image with a size147 of 512×512 pixels.

(2) The cropped images of step (1) were modified with the random contrast and brightness coefficients in the range of 0.8 to 1.2 and flipped meanwhile.

(3) Finally, the processed images of step (2) are randomly scaled with a factor between 0.8 and 1.2. Moreover, if the sizes of the processed images of step (2) are smaller than 512×512, the images are mirrored. For the purpose of the contrary, only parts of the images are cropped.

\subsection{Experimental setting}
Our experiment uses the MMSegmentation under the Open-MMLab framework, and were performed on a Windows 10 PC with Intel i7-10700K CPU and NVIDIA GeForce GTX3080 GPU. The software platform uses Cuda 11.3, PyTorch 1.11.0 and Python 3.8. In the experiment, we trained model for 300 epochs with a batch size of 8. The optimizer was Adam with an initial learning rate of 10-3, and the loss function used cross-entropy loss. The experimental results were obtained through five-fold validation. Five-fold validation could verify the effectiveness of the model. The evaluation metric used was the average of the sum.

\subsection{Evaluation metrics}
We evaluate the performance of the proposed DA-VIT model using the metrics include Pixel Accuracy (PA), Mean Intersection over Union (mIoU), Parameters (Params), and Floating-point Operations (FLOPs).

The metrics of PA and mIoU are used to assess the accuracy of the model's prediction results and reflect the segmentation performance on different groups. These are defined as follows:

\begin{equation}
	\label{deqn_ex1a}
PA = \frac{{\sum\nolimits_{i = 0}^k {{P_{ii}}} }}{{\sum\nolimits_{i = 0}^k {\sum\nolimits_{j = 0}^k {{P_{ij}}} } }}
\end{equation}

\begin{equation}
	\label{deqn_ex1a}
mIoU = \frac{1}{{k + 1}}\sum\limits_{i = 0}^k {\frac{{{P_{ii}}}}{{\sum\nolimits_{j = 0}^k {{P_{ij}}}  + \sum\nolimits_{j = 0}^k {{P_{ji}}}  - {P_{ii}}}}} 
\end{equation}
where $k$ is the total number of categories of coal maceral groups. $P_ii$ is the pixel which the true value is class $i$ and the predicted value is also class $i$. $P_ij$ is the pixel which the true value is class $i$, but the predicted value is class $j$.

The metrics of Params and FLOPs are used to evaluate the model's parameters. They can reflect the size of the model, where Params represents the number of learnable parameters in the model. FLOPs represents the computational workload required to execute the model.

\subsection{Performance of DA-VIT model}
In our experiment, we contrasted the proposed DA-VIT model with five other state-of-the-art models (Deeplabv3+\cite{chen2018encoderdecoder}, Swin-Transformer\cite{liu2021swin}, PVTv2\cite{wang2022pvt}, ConvNeXt\cite{liu2022convnet}, and VAN\cite{guo2023visual}) in the preprocessed coal microscopic image dataset from Section \uppercase\expandafter{\romannumeral3}. A. Additionally, we performed two ablation experiments to show the effectiveness and superiority of the DA-VIT model.

\subsection{Results}
\subsubsection{Comparative experiments of DA-VIT}
We use dataset A obtained from the preprocessing in Section \uppercase\expandafter{\romannumeral4}. B , Dataset B from \cite{li2022coal}, and Dataset C from \cite{tong2022coal} to test the segmentation performance of the DA-VIT model.

\begin{figure*}[t]
	\centering
	\includegraphics[width=0.9\linewidth]{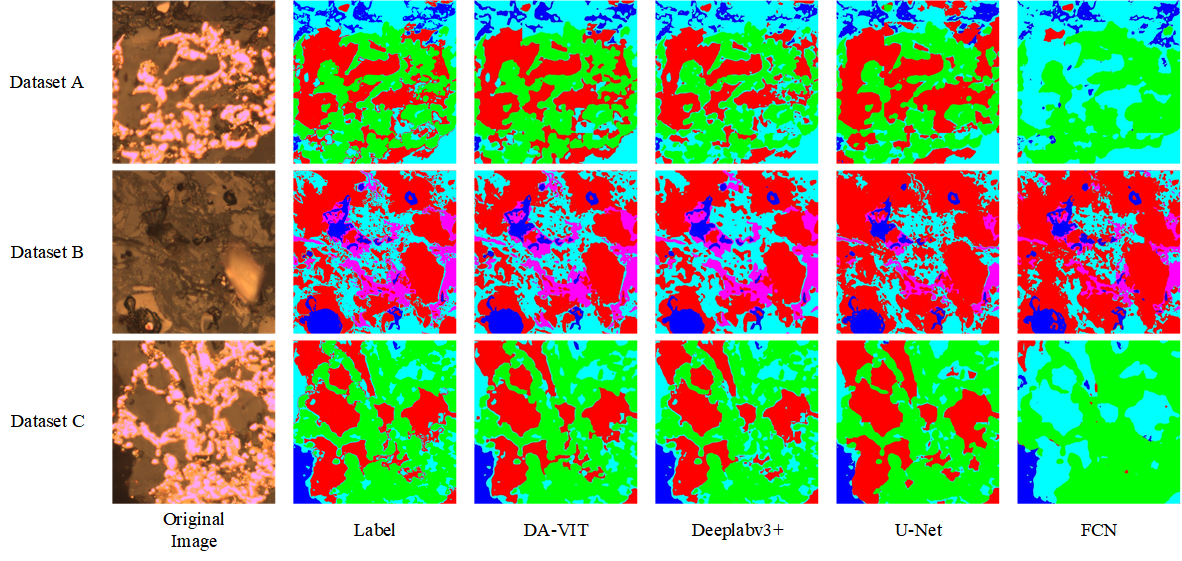}
	\caption{Comparison of DA-VIT and existing coal maceral groups segmentation model results}
	\label{fig_14}
\end{figure*}

In order to test the segmentation effect of DA-VIT under low parameters, we compare our DA-VIT model with existing deep learning used for coal maceral groups, including FCN \cite{shelhamer2017fully}, Unet \cite{ronneberger2015uneta}, Deeplabv3+ \cite{chen2018encoderdecoder}, Segnext \cite{guo2022segnext}, SETR \cite{zheng2021rethinking} and SegFormer \cite{xie2021segformer}. Fig. \ref{fig_14} indicates that DA-VIT achieves richer segmentation details and more further accurate boundary delineation. Additionally, our DA-VIT model is compared with state-of-the-art models including Swin-Transformer\cite{liu2021swin}, PVTv2\cite{wang2022pvt}, ConvNeXt\cite{liu2022convnet}, and VAN\cite{guo2023visual}. The evaluation metric results are presented in Table \ref{tab4}.

\begin{table}[!ht]
	\centering
	\caption{Comparison with segmentation models of maceral groups}
	\label{tab4}
	\begin{tabular}{cccccc}
		\hline
		&Methods & PA(\%) & mIoU(\%) & Params(M) & GFLOPs \\
		\hline
		\multirow[m]{14}{*}{\textbf{A}}&FCN & 81.41 & 48.37 & 49.49 & 197.70 \\
		&Unet & 86.22 & 52.41 & 29.06 & 202.61 \\
		&Deeplabv3+ & 88.59 & 59.63 & 43.58 & 176.24 \\
		&PVTv2-B0 & 89.37 & 60.59 & 7.53 & 23.68 \\
		&VAN-B0 & 88.95 & 60.92 & 7.97 & 25.48 \\
		&\textbf{DA-VIT-Tiny} & \textbf{90.32} & \textbf{61.70} & \textbf{4.95} & \textbf{8.99} \\
		&PVTv2-B1 & 90.07 & 61.14 & 17.74 & 31.83 \\
		&VAN-B1 & 90.20 & 62.44 & 17.60 & 34.29 \\
		&\textbf{DA-VIT-Small} & \textbf{91.17} & \textbf{62.65} & \textbf{14.74} & \textbf{18.77} \\
		&SegFormer &88.38 & 60.9 &
		48.8& 70.00\\
		&SETR &88.83 & 52.89  &87.69  & 280.3 \\
		&SegNext & 89.45 & 60.13 & 
		50.13 & 198.69 \\
		&Swin & 89.30 & 61.36 & 59.94 & 236.08 \\
		&PVTv2-B2 & 90.79 & 62.07 & 29.10 & 41.54 \\
		&VAN-B2 & 91.24 & 62.31 & 30.31 & 47.41 \\
		&ConvNeXt-T & 91.83 & 62.75 & 60.13 & 233.47 \\
		&\textbf{DA-VIT-Base} & \textbf{92.14} & \textbf{63.18} & \textbf{27.57} & \textbf{32.05} \\
		\hline
		\multirow[m]{14}{*}{\textbf{B}}
		&FCN &81.41 &48.37 &49.49 &197.70 \\
		&Unet	&86.22	&52.41	&29.06	&202.61 \\
		&Deeplabv3+	&88.59	&59.63	&43.58	&176.24\\
		&PVTv2-B0	&89.37	&60.59	&7.53	&23.68\\
		&
		VAN-B0	&88.95	&60.92	&7.97	&25.48\\
		&
		\textbf{DA-VIT-Tiny}	&\textbf{90.32}	&\textbf{61.70}	&\textbf{4.95}	&\textbf{8.99}\\
		&
		PVTv2-B1	&90.07	&61.14	&17.74	&31.83\\
		&
		VAN-B1	&90.20	&62.44	&17.60	&34.29\\
		&
		\textbf{DA-VIT-Small}	&\textbf{91.17}	&\textbf{62.65}	&\textbf{14.74}	&\textbf{18.77}\\
&SegFormer &88.38 & 60.9 &48.8& 70.00\\
& &88.83 & 52.89 &
87.69  & 280.3 \\
&SegNext & 89.45 & 60.13 & 
50.13 & 198.69 \\
		&Swin	&89.30	&61.36	&59.94	&236.08\\
		&
		PVTv2-B2	&90.79	&62.07	&29.10	&41.54\\
		&
		VAN-B2	&91.24	&62.31	&30.31	&47.41\\
		&
		ConvNeXt-T	&91.83	&62.75	&60.13	&233.47\\
		&
		\textbf{DA-VIT-Base} &	\textbf{92.14} & \textbf{63.18} &\textbf{27.57} & \textbf{32.05} \\
		\hline
		\multirow[m]{14}{*}{\textbf{C}}&
		FCN	&81.41	&48.37	&49.49	&197.70\\
		&
		Unet	&86.22&	52.41	&29.06	&202.61\\
		&
		Deeplabv3+	&88.59	&59.63	&43.58 &	176.24\\
		&
		PVTv2-B0	&89.37	&60.59	&7.53	&23.68\\
		&
		VAN-B0	&88.95	&60.92	&7.97	&25.48\\
		&
		\textbf{DA-VIT-Tiny} & \textbf{90.32} & \textbf{61.70} & \textbf{27.57}&\textbf{32.05}\\
		&
		PVTv2-B1	&90.07	&61.14	&17.74	&31.83\\
		&
		VAN-B1	&90.20	&62.44	&17.60	&34.29\\
		&
		\textbf{DA-VIT-Small} &\textbf{91.17} &\textbf{62.65} &\textbf{14.74} &\textbf{18.77}\\
		&SegFormer &88.3 & 60.9 &
		48.8& 70.00\\
		&SETR &88.83 & 52.89  &
		87.69  & 280.3 \\
		&SegNext & 89.45 & 60.13 & 
		50.13 & 198.69 \\
		&
		Swin	&89.30	&61.36	&59.94	&236.08\\
		&
		PVTv2-B2	&90.79	&62.07	&29.10	&41.54\\
		&
		VAN-B2	&91.24	&62.31	&30.31	&47.41\\
		&
		ConvNeXt-T	&91.83	&62.75	&60.13	&233.47\\
		&
		\textbf{DA-VIT-Base} & \textbf{92.14} & \textbf{63.18} & \textbf{27.57} &\textbf{32.05}\\
		\hline
	\end{tabular}
\end{table}

As shown in Table \ref{tab4}, it can be indicated that the proposed DA-VIT model exhibits significantly lower Params and FLOPs compared to existing coal maceral groups segmentation models and recent advanced semantic segmentation methods. It achieves the highest PA and mIoU of 92.14\% and 63.18\%, respectively. Both are superior to other models.

\subsubsection{Confusion matrix analysis}
To verify the segmentation accuracy of DA-VIT on each maceral group, we use a confusion matrix to evaluate the segmentation effect of the DA-VIT model. And we compare it with Deeplabv3+ \cite{chen2018encoderdecoder} which is the best coal maceral groups segmentation model. We show represent results in Table \ref{tab4} and Table \ref{tab5}. Obviously, the DA-VIT outperforms the Deeplabv3+ on various maceral groups.

\begin{table}[!ht]
	\centering
	\caption{Confusion matrix of coal maceral groups segmentation in Deeplabv3+ model}
	\label{tab5}
	\begin{tabular}{cccccc}
		\toprule
		Maceral & vitrinite & inertinite & exinite & mineral & adhesive \\
		\midrule
		vitrinite & 0.92 & 0.03 & 0 & 0.01 & 0.04 \\
		inertinite & 0.15 & 0.72 & 0.02 & 0 & 0.11 \\
		exinite & 0.05 & 0 & 0.91 & 0.01 & 0.03 \\
		mineral & 0.01 & 0.08 & 0 & 0.89 & 0.02 \\
		adhesive & 0.02 & 0.01 & 0.05 & 0 & 0.92 \\
		\bottomrule
	\end{tabular}
\end{table}

\subsubsection{Ablation studies on different modules}
To verify the effectiveness of each module in DCSA of the DA-VIT, the ablation experiments are performed on the three independent modules within DCSA using the Tiny architecture. The results of ablation experiments are listed in Table \ref{tab6} where `only LD-Conv’ means the removal of the MD-Conv and ’w/o SI-Conv’ means the removal of the single module.

\begin{table}[!ht]
	\centering
	\caption{Results of ablation experiments with different modules}
	\label{tab6}
	\begin{tabular}{ccccc}
		\toprule
		& PA(\%) & mIoU(\%) & Params(M) & GFLOPs \\
		\midrule
		only LD-Conv & 89.46 & 61.62 & 4.89 & 8.87 \\
		w/o LD-Conv & 88.91 & 61.57 & 4.87 & 8.82 \\
		w/o MD-Conv & 89.67 & 61.63 & 4.90 & 8.90 \\
		w/o SI-Conv & 89.93 & 61.68 & 4.93 & 8.95 \\
		w/o 1×1 Conv & 89.20 & 61.58 & 4.95 & 8.72 \\
		DA-VIT-Tiny & 90.32 & 61.70 & 4.95 & 8.99 \\
		\bottomrule
	\end{tabular}
\end{table}

Table \ref{tab6} illustrates that each module of the DA-VIT improves the segmentation accuracy implemented with fewer parameters and computational requirements. Specifically, SI-Conv affects the segmentation result of the coal maceral groups boundaries. LD-Conv impacts a significant on the model’s segmentation accuracy. And 1×1 Conv enhances the model’s adaptability in the channel dimension of coal microscopic images.

\subsubsection{Ablation studies on different sizes of equivalent convolutional kernels}
In Section \uppercase\expandafter{\romannumeral4}.D, we chose equivalent convolutional kernels of $K$=19. However, the model performance can be impacted by the different sizes of equivalent convolutional kernels. So, we perform the experiments about the DA-VIT model with three different scales of equivalent convolutional kernels. The smaller size equivalent convolutional kernel is $K$ =7. It consists of a 3×3 SI-Conv, a 3×3 MD-Conv, and a 3×3 LD-Conv. The larger size kernel is $K$ =25. It consists of a 5×5 SI-Conv, a 7×7 MD-Conv, and a 9×9 LD-Conv. We perform the experiments with the Tiny architecture.

\begin{table}[!ht]
	\centering
	\caption{Comparison with segmentation models of maceral groups}
	\label{tab7}
	\begin{tabular}{ccccc}
		\toprule
		K & PA(\%) & mIoU(\%) & Params(M) & GFLOPSs \\
		\midrule
		9 & 87.23 & 59.97 & 4.86 & 8.80 \\
		19 & 90.32 & 61.70 & 4.95 & 8.99 \\
		25 & 90.38 & 61.62 & 5.06 & 9.22 \\
		\bottomrule
	\end{tabular}
\end{table}

Table \ref{tab7} shows that the model accuracy decreases when the parameters are reduced by using a smaller convolutional kernel since the smaller size kernel tends to discard additional associated information in the mid-to-low layers of the DA-VIT model, resulting in the loss of long-range dependent information. Moreover, when a larger size kernel is used in the experiment, the performance of the DA-VIT model has not been significantly improved due to unnecessary parameters. The result illustrates that the deeper layers of the DA-VIT model play a more significant role in feature extraction. Furthermore, the DA-VIT model has a multi-layer down-sampling structure, which makes an overly large equivalent convolutional kernel becomes redundant in the deeper layers of DA-VIT, it cannot improve the accuracy of DA-VIT.

\subsubsection{Analysis of the DA-VIT segmentation accuracy with dataset expansion through the IoT}
\begin{figure}[!ht]
	\centering
	\includegraphics[width=\linewidth]{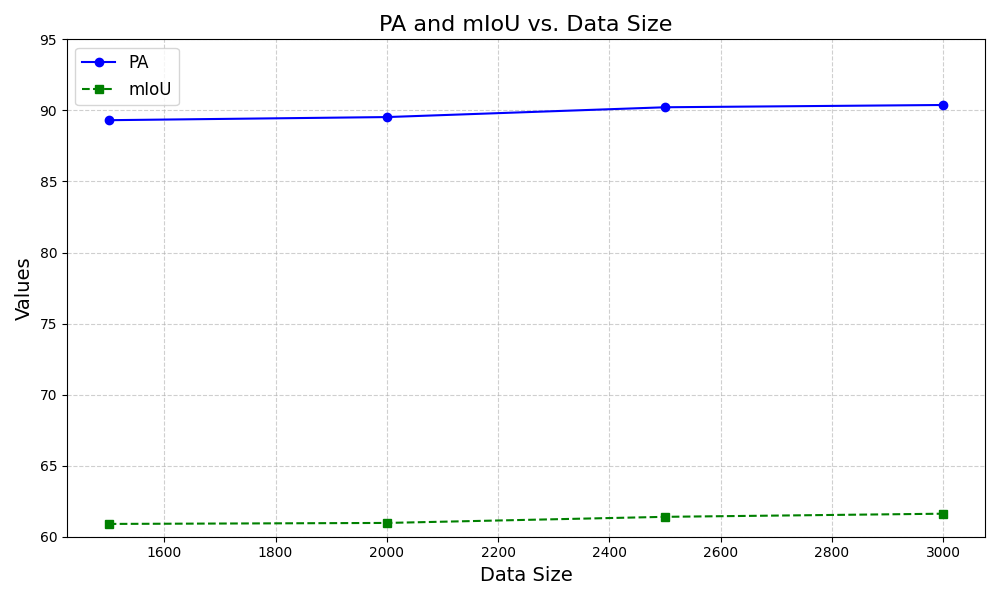}
	\caption{Changes in evaluation indicators as a function of data volume}
	\label{fig_23}
\end{figure}

Both table \ref{tab:tab10} and \ref{fig_23} illustrate that the segmentation results of the DA-VIT model improve as we continuously increase the dataset size through the IoT. As the dataset size grew, the PA and mIoU increased 0.57\% and 0.31\% in average respectively, which demonstrates the method of expanding the dataset based on a multi-expert evaluate the predicted data within the IoT approach, significantly enhances the DA-VIT model's ability to extract features of coal maceral groups.

\begin{table}[!ht]
	\centering
	\caption{Dataset Size and DA-VIT Evaluation Metric Results}
	\label{tab:tab10}
	\begin{tabularx}{\linewidth}{c>{\centering\arraybackslash}X>{\centering\arraybackslash}X>{\centering\arraybackslash}X>{\centering\arraybackslash}X}
		\hline
		Dataset Size & PA(\%) & mIoU(\%) & Params(M) & GFLOPSs \\
		\hline
		1500 & 89.31 & 60.90 & \multirow[m]{4}{*}{5.06} & \multirow[m]{4}{*}{9.22} \\
		2000 & 89.53 & 60.97 &  &  \\
		2500 & 90.22 & 61.40 &  &  \\
		3000 & 90.38 & 61.62 &  &  \\
		\hline
	\end{tabularx}
\end{table}

\subsubsection{Influence of dilation rate $r$ in dilation convolution on the segmentation effect of DA-VIT models}
Table \ref{tab9} demonstrates the effect of different r on model performance. When $r$ is less than 2, PA and mIoU increased by 0.94\% and 0.69\% on average, respectively. This indicates that the model acquires more microcomponent characterization information. When $r$ is 2, PA and mIoU reached the maximum values of 92.14\% and 63.18\%, respectively. At this time, the model achieves the best segmentation performance. However, when $r$ was larger than 2 time, PA and mIoU gradually decreased. Although the expansion of $r$ can increase the receptive field to extract more feature information for the model, it also leads to the loss of more detailed features of the microscopic components. Therefore, the model has the best segmentation performance when $r$ is 2.

\begin{table}[!ht]
	\centering
	\caption{Effect of dilation rate $r$ on the performance of the DA-VIT model}
	\label{tab9}
	\begin{tabular}{ccc}
		\toprule
		$r$ & PA(\%) & mIoU(\%)  \\
		\midrule
		0 & 90.23 & 61.80 \\
		1 & 91.20 & 62.49  \\
		2 & 92.14 & 63.18  \\
		3 & 90.38 & 61.62 \\
		4 & 89.63 & 60.88 \\
		\bottomrule
	\end{tabular}
\end{table}

\section{Conclusion}
This paper innovatively using IoT to solve the problems of limited, singular sources of training data samples and weak generalization ability in deep learning-based methods for coal maceral group images training and proposes a novel DCSA mechanism and a lightweight DA-VIT segmentation model based on DCSA to significantly improve computational efficiency. This paper primarily integrates DA-VIT with IoT and constructs a parallel DA-VIT network structure within the IoT. We validate the proposed method and analyze it using multiple datasets in Section \uppercase\expandafter{\romannumeral4}.F . The proposed method effectively solves the problems of high time costs, slow inference speed, and low segmentation accuracy in the closed-loop activities of "data acquisition," "data processing," and "result transmission" in coal microanalysis technology.

(1) DCSA can significantly reduce the parameter of the convolutional self-attention. DCSA combines the attention mechanism with the dilated convolution and performs a multi-scale decomposition to the large-size kernels. The size of parameters can be reduced by 81.18\% with DCSA. We construct the coal maceral group industrial IoT architecture, carried out image pre-processing and normalization at each terminal, and finally obtained different types of coal maceral group images from multiple terminals and constituted datasets. 

(2) DA-VIT can improve the segmentation effect of coal maceral groups. DA-VIT uses the Swin-Transformer framework, and the DCSA mechanism of DA-VIT captures two-dimensional spatial information and long-range dependencies between global pixels of coal microscopic images to improve accuracy. The experimental results demonstrate that the results of PA are 90.32\%, 91.17\%, and 92.14\%, respectively, compared with the MIoU of 61.70\%, 62.65\%, and 63.18\% under three different structures of DA-VIT.

(3) While ensuring accuracy, the overall parameters of DA-VIT are effectively reduced due to the reduced parameters in DCSA. In our experiment, under three different structures of DA-VIT, params are 4.95M, 14.74M, and 27.57M, respectively. The calculated amount index FLOPs are 8.99G, 18.77G, and 32.05G, respectively. Both params and FLOPs are lower than other models with the same segmentation accuracy.

(4) We have demonstrated the effectiveness of DA-VIT for segmentation of coal rock microcomponent groups. In addition, we have also thought about the adaptability of the DA-VIT model proposed in this paper to other materials. Due to the coal rock microgroup images, different microgroups have obvious boundaries, more concentrated distribution, and multiple microgroups present blocky and other characteristics. The method proposed in this paper should also be applicable to other materials with similar characteristics. However, it is necessary to preprocess the material images and labeled images according to the method in Section IV.B to construct the dataset and train the DA-VIT model. In the future, this paper will add other material images to explore the generalization of the IoT-based DA-VIT model for parallel structures.

Besides, we have applied successfully in practical production at Angang Steel Group Limited in China.

\section*{Acknowledgments}
This work was jointly supported by the project of promoting research paradigm reform , empowering disciplinary leap through artificial intelligence from Shanghai municipal education commission (Grant numbers 2024AI011) and the National Natural Science Foundation of China (Grant numbers 12104289) and was gratefully acknowledged.

\bibliographystyle{ieeetr}
\bibliography{references}

\begin{IEEEbiography}[{\includegraphics[width=1in,height=1.25in,clip,keepaspectratio]{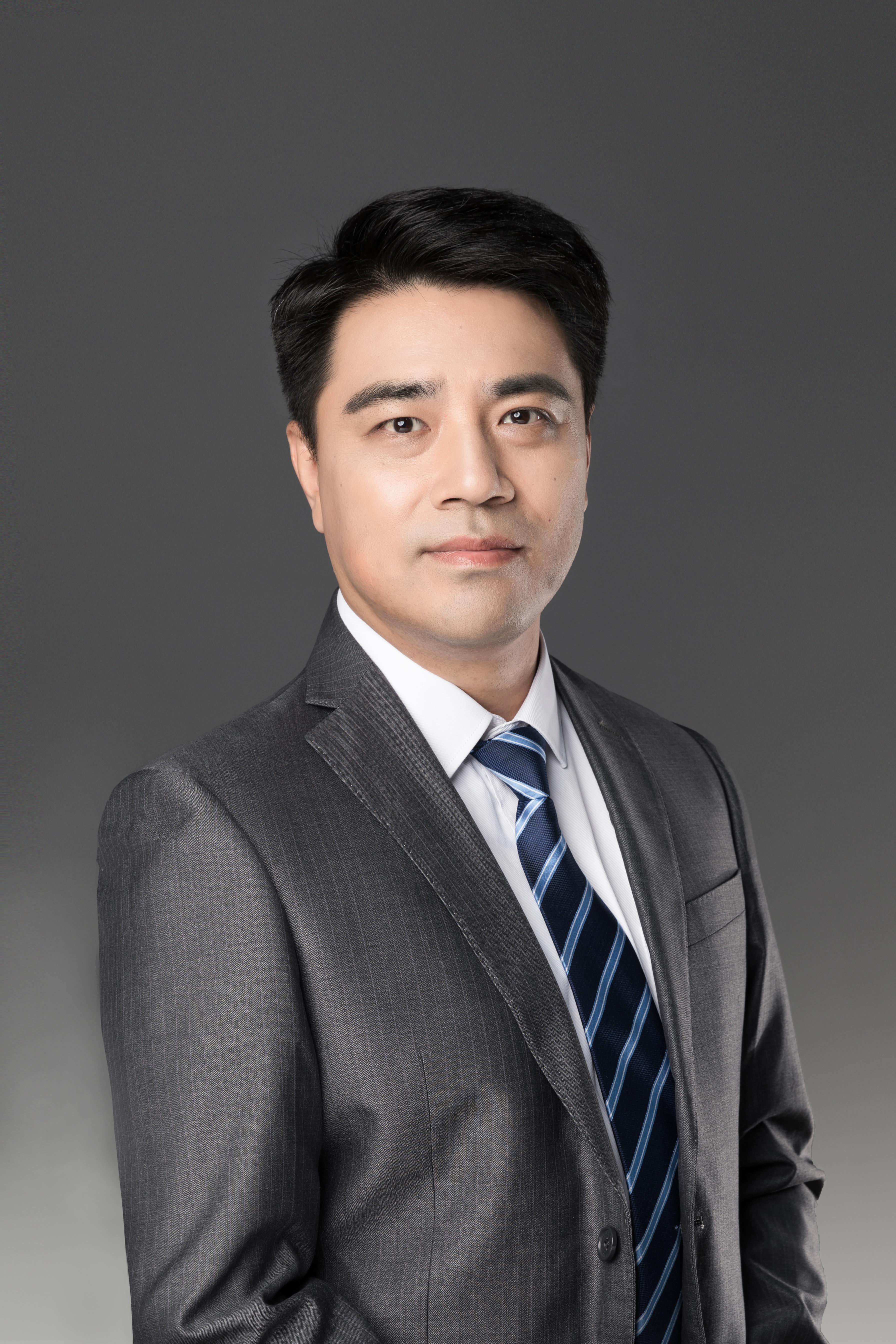}}]{Zhenghao Xi (Member, IEEE)}
	received the B.E. degree from the South Central University for Nationalities, Wuhan, China, in 2003, the M.E. degree from the University of Science and Technology Liaoning, Anshan, China, in 2008, and the Ph.D. degree from the University of Science and Technology Beijing, Beijing, China, in 2015. 
	
	He is currently a Professor with the Shanghai University of Engineering Science, Shanghai, China, and also the Director of the Department of Automation. His research interests include computer vision, robotics, and intelligent system.
\end{IEEEbiography}

\begin{IEEEbiography}[{\includegraphics[width=1in,height=1.25in,clip,keepaspectratio]{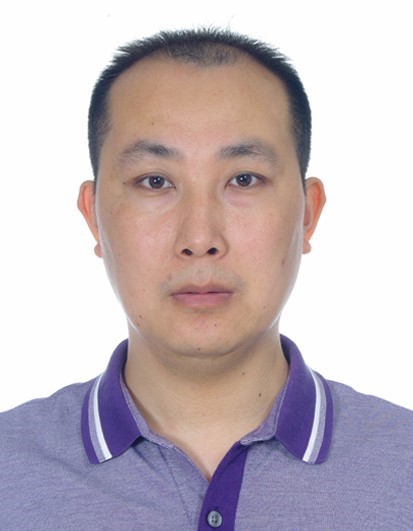}}]{Zhengnan Lv}
	was born in LiaoNing, China, in 1980. From 1999 to 2003, he studied in Jilin University and received his bachelor's degree in 2003. From 2003 to 2006, he studied in Jilin University and received his Master’s degree in 2006. CHis current main research interests are	Internet of Things, artificial intelligence, wireless mobile communication, and network security.
\end{IEEEbiography}

\begin{IEEEbiography}[{\includegraphics[width=1in,height=1.25in,clip,keepaspectratio]{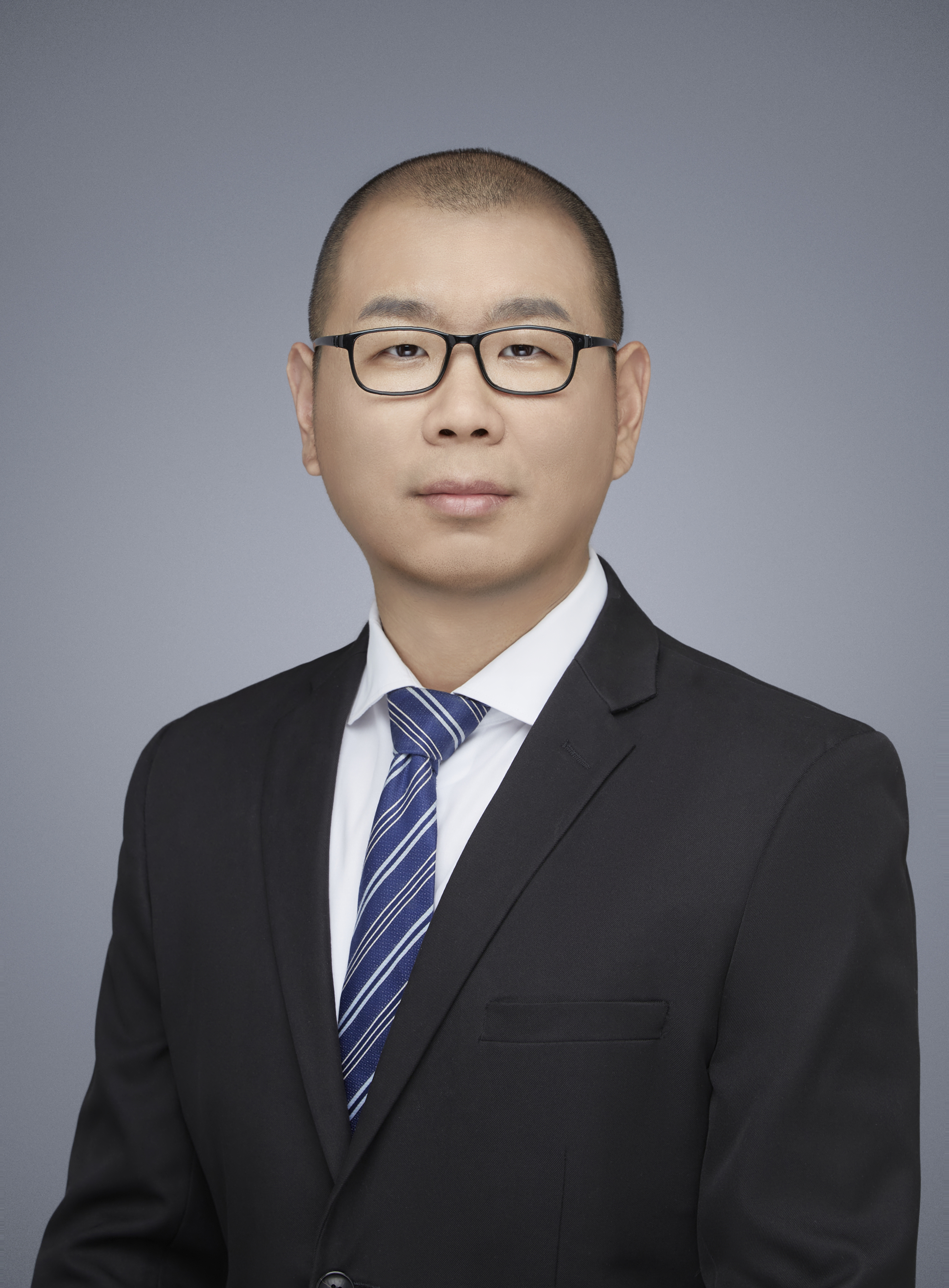}}]{Yang Zheng (Member, IEEE)}
	received Ph.D. degree from the University of Science and Technology Beijing, China, in 2018.
	
	He is currently an associate researcher at the Institute of Automation, Chinese Academy of Sciences. His research interests include computer vision, speech recognition, and other related fields.
\end{IEEEbiography}

\begin{IEEEbiography}[{\includegraphics[width=1in,height=1.25in,clip,keepaspectratio]{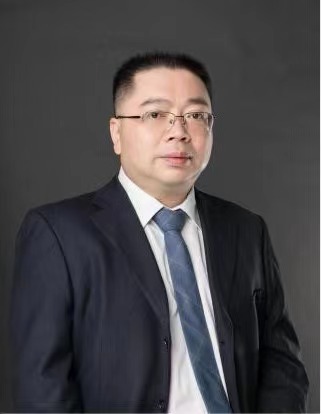}}]{Xiang Liu (Member, IEEE)}
	received Ph.D. degree from Fudan University, MEng degree from Jiangsu University and BSc degree from Nanjing Normal University. 
	He is currently the professor and director of Computer Department with School of Electronic and Electric Engineering of Shanghai University of Science Engineering, Shanghai, China. His current research interests include computer vision and pattern recognition.
\end{IEEEbiography}

\begin{IEEEbiography}[{\includegraphics[width=1in,height=1.25in,clip,keepaspectratio]{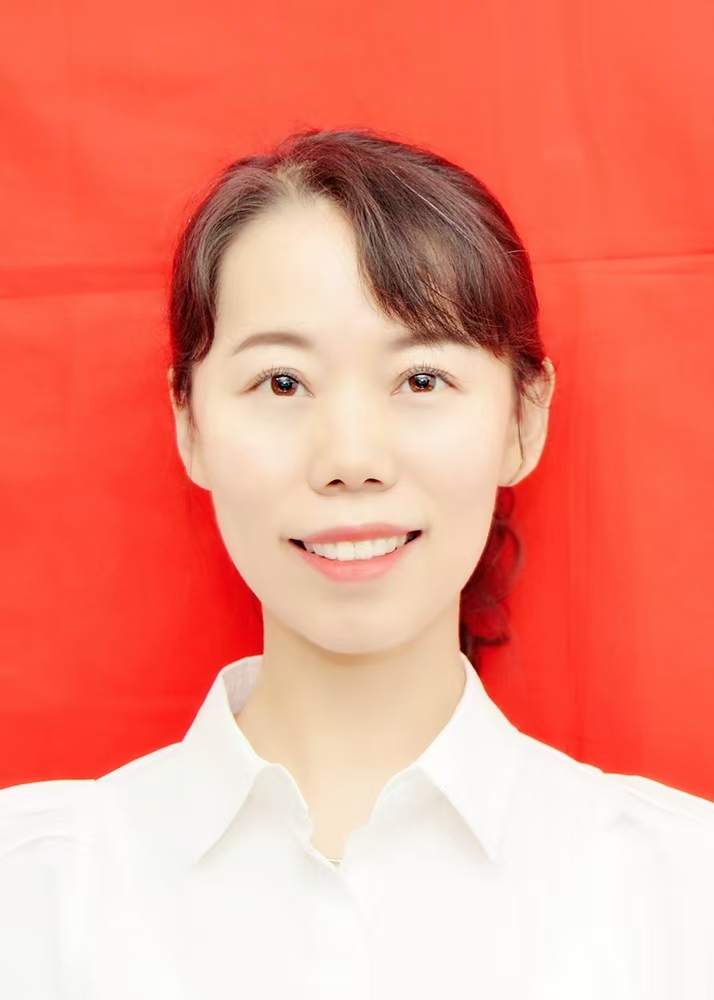}}]{ Zhuang Yu }
received the M.E. degree from the University of Science and Technology Liaoning, Anshan, China, in 2008.She is currently an coal analyst at the department of manufacture management of angang steel co.,ltd. Her research interests include the analysis of coal.
\end{IEEEbiography}

\begin{IEEEbiography}[{\includegraphics[width=1in,height=1.25in,clip,keepaspectratio]{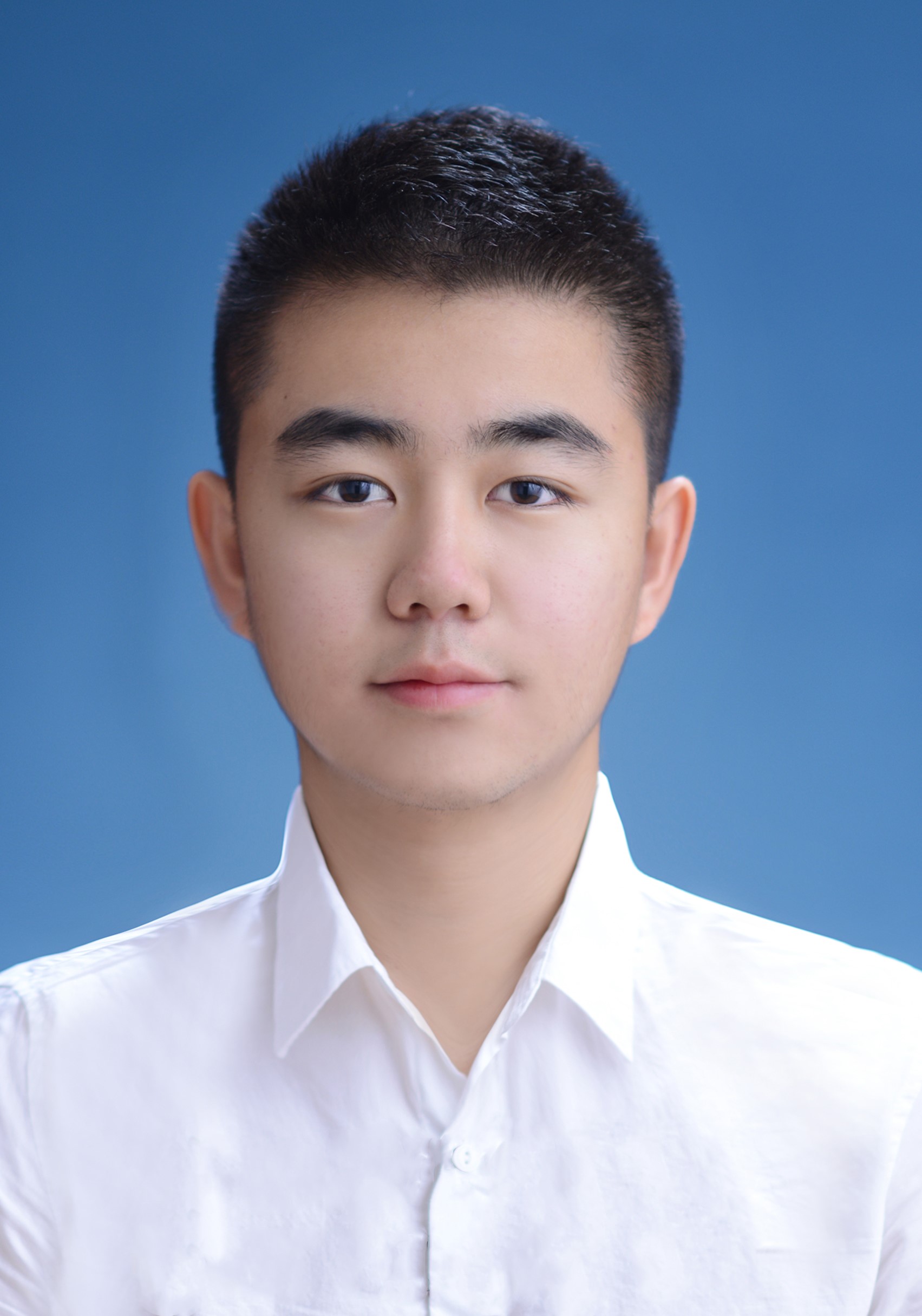}}]{Junran Chen}
	(Student Member, IEEE) received th B.E. degree from Shanghai University of Engineering Science, Shanghai, China, in 2022. He is currently pursuing the M.S. degree with the Shanghai University of Engineering Science, Shanghai, China.
	
	His research interests are image processing, image segmentation and image recognition.
\end{IEEEbiography}

\begin{IEEEbiography}[{\includegraphics[width=1in,height=1.25in,clip,keepaspectratio]{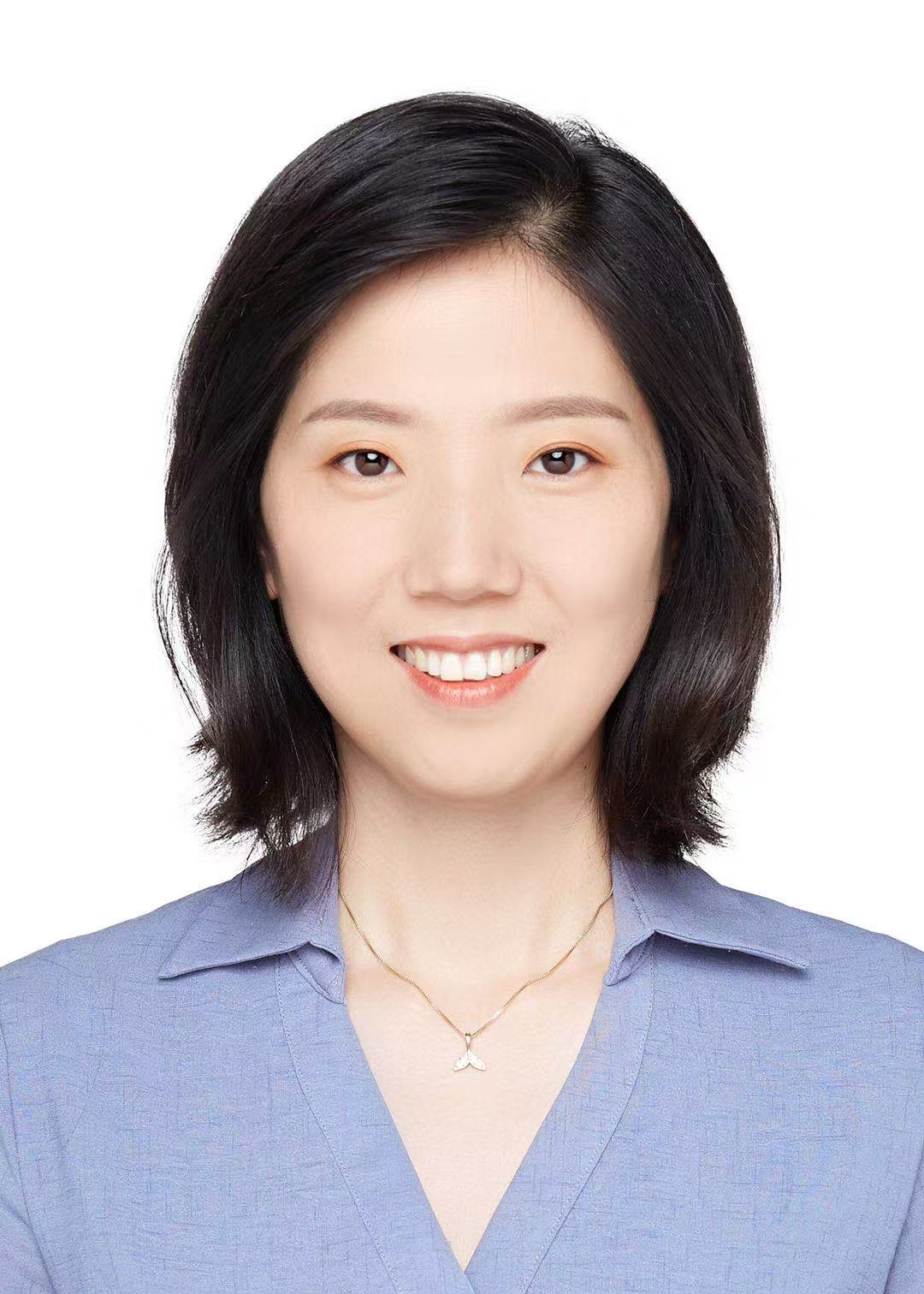}}]{Jing Hu}
Jing Hu received the master’s degree in information and telecommunications engineering from Xidian University, Xi’an, China, in 2006.Her current main research
directions are wireless communication, digital twins, and artificial intelligence.
\end{IEEEbiography}

\begin{IEEEbiography}[{\includegraphics[width=1in,height=1.25in,clip,keepaspectratio]{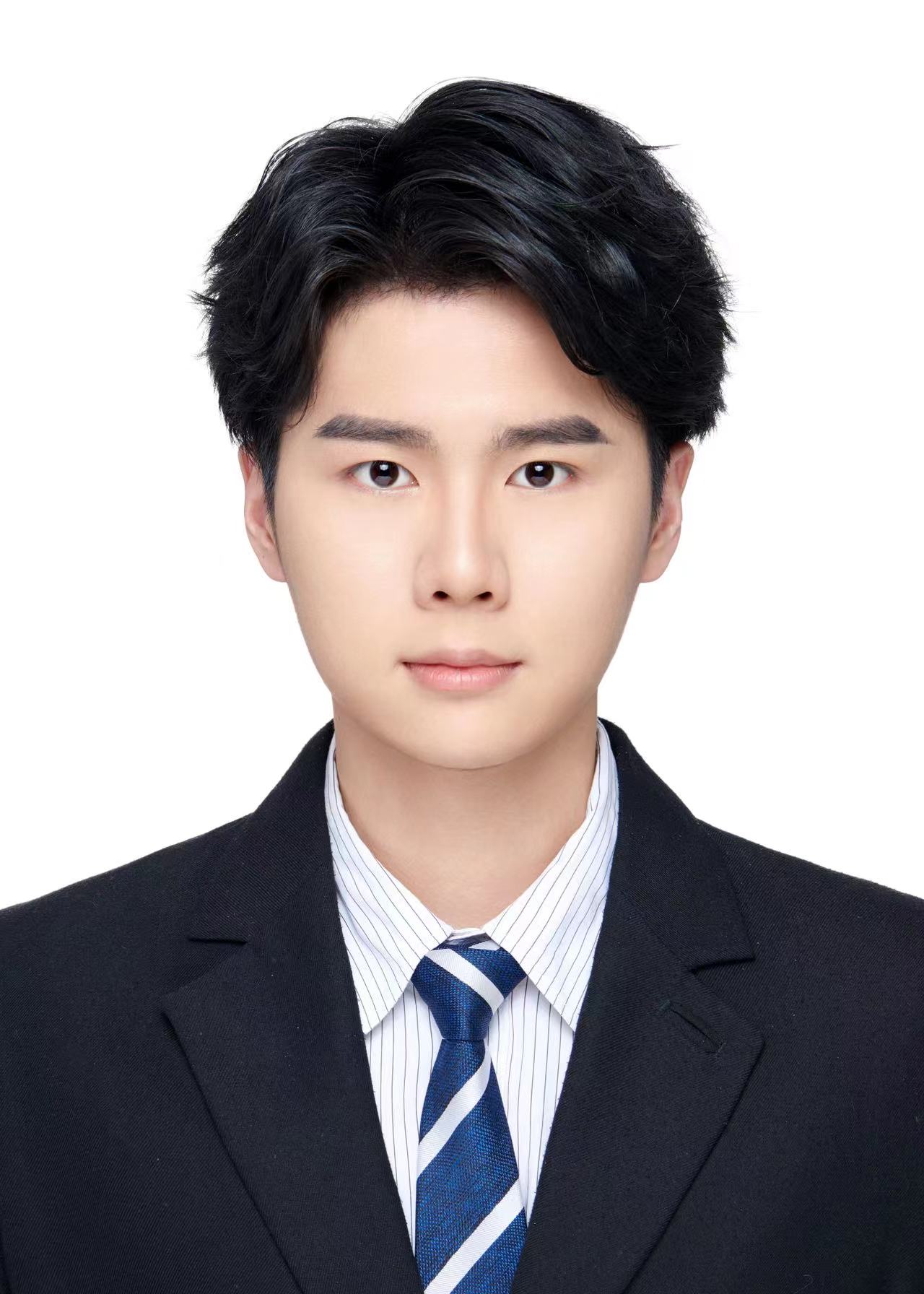}}]{Yaqi Liu}
	received the B.E. degree from Shanghai University of Engineering Science, Shanghai, China, in 2024, where he is currently working toward the M.S. degree. His research interests include computer vision and robotics.
\end{IEEEbiography}

\end{document}